# Tagging French –
# comparing a statistical and a constraint-based method

Jean-Pierre Chanod and Pasi Tapanainen
Rank Xerox Research Centre, Grenoble Laboratory
6, chemin de Maupertuis, 38240 Meylan, France
Jean.Pierre.Chanod,Pasi.Tapanainen@xerox.fr

## Abstract

In this paper we compare two competing approaches to part-of-speech tagging, statistical and constraint-based disambiguation, using French as our test language. We imposed a time limit on our experiment: the amount of time spent on the design of our constraint system was about the same as the time we used to train and test the easy-to-implement statistical model. We describe the two systems and compare the results. The accuracy of the statistical method is reasonably good, comparable to taggers for English. But the constraint-based tagger seems to be superior even with the limited time we allowed ourselves for rule development.

## 1  Overview

In this paper[1] we compare two competing approaches to part-of-speech tagging, statistical and constraint-based disambiguation, using French as our test language. The process of tagging consists of three stages: tokenisation, morphological analysis and disambiguation. The two taggers include the same tokeniser and morphological analyser.

The tokeniser uses a finite-state transducer that reads the input and outputs a token whenever it has read far enough to be sure that a token is detected. The morphological analyser contains a transducer lexicon. It produces all the legitimate tags for words that appear in the lexicon. If a word is not in the lexicon, a guesser is consulted. The guesser employs another finite-state transducer. It reads a token and prints out a set of tags depending on prefixes, inflectional information and productive endings that it finds.

We make even more use of transducers in the constraint-based tagger. The tagger reads one sentence at a time, a string of words and alternative tags, feeds them to the grammatical transducers that remove all but one alternative tag from all the words on the basis of contextual information.

If all the transducers described above (tokeniser, morphological analyser and disambiguator) could be composed together, we would get one single transducer that transforms a raw input text to a fully disambiguated output.

The statistical method contains the same tokeniser and morphological analyser. The disambiguation method is a conventional one: a hidden Markov model.

## 2  Morphological analysis and guessing

The morphological analyser is based on a lexical transducer (Karttunen et al., 1992). The transducer maps each inflected surface form of a word to its canonical lexical form followed by the appropriate morphological tags.

Words not found in the lexicon are analysed by a separate finite-state transducer, the guesser. We developed a simple, extremely compact and efficient guesser for French. It is based on the general assumption that neologisms and uncommon words tend to follow regular inflectional patterns. The guesser is thus based on productive endings (like *ment* for adverbs, *ible* for adjectives, *er* for verbs). A given ending may of course point to various categories, e.g. *er* identifies nouns as well as verbs due to possible borrowings from English.

## 3  The statistical model

We use the Xerox part-of-speech tagger (Cutting *et al.*, 1992), a statistical tagger made at the Xerox Palo Alto Research Center.

### 3.1  Training

The Xerox tagger is claimed (Cutting *et al.*, 1992) to be adaptable and easily trained; only a lexicon and suitable amount of untagged text is required. A new language-specific tagger can therefore be built with a minimal amount of work. We started our project by doing so. We took our lexicon with the new tagset, a corpus of French text, and trained the tagger. We ran the tagger on another text and counted the errors. The result was not good; 13 % of the words were tagged incorrectly.

---

[1] There is a longer version (17 pages) of this paper in (Chanod and Tapanainen, 1994)

The tagger does not require a tagged corpus for training, but two types of biases can be set to tell the tagger what is correct and what is not: symbol biases and transition biases. The symbol biases describe what is likely in a given ambiguity class. They represent kinds of lexical probabilities. The transition biases describe the likelihood of various tag pairs occurring in succession. The biases serve as initial values before training.

We spent approximately one man-month writing biases and tuning the tagger. Our training corpus was rather small, because the training had to be repeated frequently. When it seemed that the results could not be further improved, we tested the tagger on a new corpus. The eventual result was that 96.8 % of the words in the corpus were tagged correctly. This result is about the same as for statistical taggers of English.

### 3.2 Modifying the biases

A 4 % error rate is not generally considered a negative result for a statistical tagger, but some of the errors are serious. For example, a sequence of *determiner...noun...noun/verb...preposition* is frequently disambiguated in the wrong way, e.g. *Le train part à cinq heures* (*The train leaves at 5 o'clock*). The word *part* is ambiguous between a noun and a verb (singular, third person), and it is disambiguated incorrectly. The tagger seems to prefer the noun reading between a singular noun and a preposition.

One way to resolve this is to write new biases. We added two new ones. The first one says that a singular noun is not likely to be followed by a noun (this is not always true but we could call this a tendency). The second states that a singular noun is likely to be followed by a singular, third-person verb. The result was that the problematic sentence was disambiguated correctly, but the changes had a bad side effect. The overall error rate of the tagger increased by over 50 %. This illustrates how difficult it is to write good biases. Getting a correct result for a particular sentence does not necessarily increase the overall success rate.

## 4 The constraint-based model

### 4.1 A two-level model for tagging

In the constraint-based tagger, the rules are represented as finite-state transducers. The transducers are composed with the sentence in a sequence. Each transducer may remove, or in principle it may also change, one or more readings of the words. After all the transducers have been applied, each word in the sentence has only one analysis.

Our constraint-based tagger is based on techniques that were originally developed for morphological analysis. The disambiguation rules are similar to phonological rewrite rules (Kaplan and Kay, 1994), and the parsing algorithm is similar to the algorithm for combining the morphological rules with the lexicon (Karttunen, 1994).

The tagger has a close relative in (Koskenniemi, 1990; Koskenniemi et al., 1992; Voutilainen and Tapanainen, 1993) where the rules are represented as finite-state machines that are conceptually intersected with each other. In this tagger the disambiguation rules are applied in the same manner as the morphological rules in (Koskenniemi, 1983). Another relative is represented in (Roche and Schabes, 1994) which uses a single finite-state transducer to transform one tag into another. A constraint-based system is also presented in (Karlsson, 1990; Karlsson et al., 1995). Related work using finite-state machines has been done using local grammars (Roche, 1992; Silberztein, 1993; Laporte, 1994).

### 4.2 Writing the rules

#### 4.2.1 Studying ambiguities

One quick experiment that motivated the building of the constraint-based model was the following: we took a million words of newspaper text and ranked ambiguous words by frequency. We found that a very limited set of word forms covers a large part of the total ambiguity. The 16 most frequent ambiguous word forms[2] account for 50 % of all ambiguity. Two thirds of the ambiguity are due to the 97 most frequent ambiguous words[3].

Another interesting observation is that the most frequent ambiguous words are usually words which are in general corpus-independent, i.e. words that belong to closed classes (determiners, prepositions, pronouns, conjunctions), auxiliaries, common adverbials or common verbs, like *faire* (to do, to make). The first corpus-specific word is in the 41st position.

#### 4.2.2 Principled rules

For the most frequent ambiguous word forms, one may safely define principled contextual restrictions to resolve ambiguities. This is in particular the case for clitic/determiner ambiguities attached to words like *le* or *la*. Our rule says that clitic pronouns are attached to a verb and determiners to a noun with possibly an unrestricted number of premodifiers. This is a good starting point although some ambiguity remains as in *la*

---

[2] Namely *de, la, le, les, des, en, du, un, a, dans, une, pas, est, plus, Le, son*

[3] A similar experiment shows that in the Brown corpus 63 word forms cover 50 % of all the ambiguity, and two thirds of the ambiguity is covered by 220 word forms.

*place*, which can be read as a determiner-noun or clitic-verb sequence.

Some of the very frequent words have categories that are rare, for instance the auxiliary forms *a* and *est* can also be nouns and the pronoun *cela* is also a very rare verb form. In such a case, we restrict the use of the rarest categories to contexts where the most frequent reading is not at all possible, otherwise the most frequent reading is preferred. For instance, the word *avions* may be a noun or an auxiliary verb. We prefer the noun reading and accept the verb reading only when the first-person pronoun *nous* appears in the left context, e.g. as in *nous ne les avions pas* (we did not have them).

This means that the tagger errs only when a rare reading should be chosen in a context where the most common reading is still acceptable. This may never actually occur, depending on how accurate the contextual restrictions are. It can even be the case that discarding the rare readings would not induce a detectable loss in accuracy, e.g. in the conflict between *cela* as a pronoun and as a verb. The latter is a rarely used tense of a rather literary verb.

The principled rules do not require any tagged corpus, and should be thus corpus-independent. The rules are based on a short list of extremely common words (fewer than 100 words).

### 4.2.3 Heuristics

The rules described above are certainly not sufficient to provide full disambiguation, even if one considers only the most ambiguous word forms. We need more rules for cases that the principled rules do not disambiguate.

Some ambiguity is extremely difficult to resolve using the information available. A very problematic case is the word *des*, which can either be a determiner, *Jean mange des pommes* (Jean eats apples) or an amalgamated preposition-determiner, as in *Jean aime le bruit des vagues* (Jean likes the sound of waves).

Proper treatment of such an ambiguity would require verb subcategorisation and a description of complex coordinations of noun and prepositional phrases. This goes beyond the scope of both the statistical and the constraint-based taggers. For such cases we introduce ad-hoc heuristics. Some are quite reasonable, e.g. the determiner reading of *des* is preferred at the begining of a sentence. Some are more or less arguable, e.g. the prepositional reading is preferred after a noun.

One may identify various contexts in which either the noun or the adjective can be preferred. Such contextual restrictions (Chanod, 1993) are not always true, but may be considered reasonable for resolving the ambiguity. For instance, in the case of two successive noun/adjective ambiguities like *le franc fort* (*the strong franc* or *the frank fort*), we favour the noun–adjective sequence except when the first word is a common prenominal adjective such as *bon, petit, grand, premier, ...* as in *le petit fort* (*the small fort*) or even *le bon petit* (*the good little one*).

### 4.2.4 Non-contextual rules

Our heuristics do not resolve all the ambiguity. To obtain the fully unambiguous result we make use of non-contextual heuristics. The non-contextual rules may be thought of as lexical probabilities. We guess what the most probable tag is in the remaining ambiguities. For instance, preposition is preferred to adjective, pronoun is preferred to past participle, etc. The rules are obviously not very reliable, but they are needed only when the previous rules fail to fully disambiguate.

### 4.2.5 Current rules

The current system contains 75 rules, consisting of:

- 39 reliable contextual rules dealing mostly with frequent ambiguous words.
- 25 rules describing heuristics with various degrees of linguistic generality.
- 11 non-contextual rules for the remaining ambiguities.

The rules were constructed in less than one month, on the basis of 50 newspaper sentences. All the rules are currently represented by 11 transducers.

## 5 The results

### 5.1 Test A

For evaluation, we used a corpus totally unrelated to the development corpus. It contains 255 sentences (5752 words) randomly selected from a corpus of economic reports. About 54 % of the words are ambiguous. The text is first tagged manually without using the disambiguators, and the output of the tagger is then compared to the hand-tagged result.

If we apply all the rules, we get a fully disambiguated result with an error rate of only 1.3 %. This error rate is much lower than the one we get using the hidden Markov model (3.2 %). See Figure 1.

We can also restrict the tagger to using only the most reliable rules. Only 10 words lose the correct tag when almost 2000 out of 3085 ambiguous words are disambiguated. Among the remaining 1136 ambiguous words about 25 % of the ambiguity is due to determiner/preposition ambiguities (words like *du* and *des*), 30 % are adjective/noun ambiguities and 18 % are noun/verb ambiguities.

If we use both the principled and heuristic rules, the error rate is 0.52 % while 423 words remain ambiguous. The non-contextual rules that eliminate the remaining 423 ambiguities produce an additional 43 errors. Overall, 98.7 % of the words receive the correct tag.

|  | error rate (correctness) | remaining ambiguity | tag / word |
|---|---|---|---|
| Lexicon + Guesser | 0.03 % (99.97 %) | 54 % | 1.64 |
| Hidden Markov model | 3.2 % (96.8 %) | 0 % | 1.00 |
| Principled rules | 0.17 % (99.83 %) | 20 % | 1.24 |
| Principled and heuristic rules | 0.52 % (99.48 %) | 7 % | 1.09 |
| All the rules | 1.3 % (98.7 %) | 0 % | 1.00 |

Figure 1: The result in the test sample

### 5.2 Test B

We also tested the taggers with more difficult text. The 12 000 word sample of newspaper text has typos and proper names[4] that match an existing word in the lexicon. Problems of the latter type are relatively rare but this sample was exceptional. Altogether the lexicon mismatches produced 0.5 % errors to the input of the taggers. The results are shown in Figure 2. This text also seems to be generally more difficult to parse than the first one.

### 5.3 Combination of the taggers

We also tried combining the taggers, using first the rules and then the statistics (a similar approach was also used in (Tapanainen and Voutilainen, 1994)). We evaluated the results obtained by the following sequence of operations:

1) Running the constraint-based tagger without the final, non-contextual rules.

2) Using the statistical disambiguator independently. We select the tag proposed by the statistical disambiguator if it is not removed during step 1.

3) Solving the remaining ambiguities by running the final non-contextual rules of the constraint-based tagger. This last step ensures that one gets a fully disambiguated text. Actually only about 0.5 % of words were not fully disambiguated after step 2.

We used the test sample B. After the first step, 1400 words out of 12 000 remain ambiguous. The process of combining the three steps described above eventually leads to more errors than running the constraint-based tagger alone. The statistical tagger introduces 220 errors on the 1400 words that remain ambiguous after step 1. In comparison, the final set of non-contextual rules introduces around 150 errors on the same set of 1400 words. We did not expect this result. One possible explanation for the superior performance of the final non-contextual rules is that they are meant to apply after the previous rules failed to disambiguate the word. This is in itself useful

[4]like *Bats, Botta, Démis, Ferrasse, Hersant*, ...

information. The final heuristics favour tags that have survived all conditions that restrict their use. For instance, the contextual rules define various contexts where the preposition tag for *des* is preferred. Therefore, the final heuristics favours the determiner reading for *des*.

## 6 Analysis of errors

### 6.1 Errors of principled and heuristic rules

Let us now consider what kind of errors the constraint-based tagger produced. We do not deal with errors produced by the last set of rules, the non-contextual rules, because it is already known that they are not very accurate. To make the tagger better, they should be replaced by writing more accurate heuristic rules.

We divide the errors into three categories: (1) errors due to multi-word expressions, (2) errors that should/could be resolved and (3) errors that are hard to resolve by using the information that is available.

The first group (15 errors), the multi-word expressions, are difficult for the syntax-based rules because in many cases the expression does not follow any conventional syntactic structure, or the structure may be very rare. In multi-word expressions some words also have categories that may not appear anywhere else. The best way to handle them is to lexicalise these expressions. When a possible expression is recognised we can either collapse it into one unit or leave it otherwise intact except that the most "likely" interpretation is marked.

The biggest group (41 errors) contains errors that could have been resolved correctly but were not. The reason for this is obvious: only a relatively small amount of time was allowed for writing the rules. In addition, the rules were constructed on the basis of a rather small set of example sentences. Therefore, it would be very surprising if such errors did not appear in the test sample taken from a different source. The errors are the following:

- The biggest subgroup has 19 errors that require modifications to existing rules. Our rules were meant to handle such cases but fail

|                              | error rate (correctness) | remaining ambiguity | tag / word |
|------------------------------|--------------------------|---------------------|------------|
| Lexicon + Guesser            | 0.5 % (99.5 %)           | 48 %                | 1.59       |
| Hidden Markov model          | 5.0 % (95.0 %)           | 0 %                 | 1.00       |
| Principled rules             | 0.8 % (99.2 %)           | 23 %                | 1.29       |
| Principled and heuristic rules | 1.3 % (98.7 %)         | 12 %                | 1.14       |
| All the rules                | 2.5 % (97.5 %)           | 0 %                 | 1.00       |

Figure 2: The result in a difficult test sample with many lexicon mismatches

to do so correctly in some sentences. Often only a minor correction is needed.

- Some syntactic constructions, or word sequences, were omitted. This caused 7 errors which could easily be avoided by writing more rules. For instance, a construction like "preposition + clitic + finite verb" was not forbidden. The phrase *à l'est* was analysed in this way while the correct analysis is "preposition + determiner + noun".

- Sometimes a little bit of extra lexical information is required. Six errors would require more information or the kind of refinement in the tag inventory that would not have been appropriate for the statistical tagger.

- Nine errors could be avoided by refining existing heuristics, especially by taking into account exceptions for specific words like *point*, *pendant* and *devant*.

The remaining errors (28 errors) constitute the price we pay for using the heuristics. Removing the rules which fail would cause a lot of ambiguity to remain. The errors are the following:

- Fifteen errors are due to the heuristics for *de* and *des*. There is little room for improvement at this level of description (see Chapter 4.2.3). However, the current, simple heuristics fully disambiguate 850 instances of *de* and *des* out of 914 i.e. 92 % of all the occurrences were parsed with less than a 2 % error rate.

- Six errors involve noun–adjective ambiguities that are difficult to solve, for instance, in a subject or object predicate position.

- Seven errors seem to be beyond reach for various reasons: long coordination, rare constructions, etc. An example is *les boîtes* (the boxes) where *les* is wrongly tagged in the test sample because the noun form is misspelled as *boites*, which is identified only as a verb by the lexicon.

### 6.2 Difference between the taggers

We also investigated how the errors compare between the two taggers. Here we used the fully disambiguated outputs of the taggers. The errors belong mainly to three classes:

- Some errors appear predominantly with the statistical tagger and almost never with the constraint-based tagger. This is particularly the case with the ambiguity between past participles and adjectives.

- Some errors are common to both taggers, the constraint-based tagger generally being more accurate (often with a ratio of 1 to 2). These errors cover ambiguities that are known to be difficult to handle in general, such as the already mentioned determiner/preposition ambiguity.

- Finally, there are errors that are specific to the constraint-based tagger. They are often related to errors that could be corrected with some extra work. They are relatively infrequent, thus the global accuracy of the constraint-based tagger remains higher.

The first two classes of errors are generally difficult to correct. The easiest way to improve the constraint-based tagger is to concentrate on the final class. As we mentioned earlier, it is not very easy to change the behaviour of the statistical tagger in one place without some side-effects elsewhere. This means that the errors of the first class are probably easiest to resolve by means other than statistics.

The first class is quite annoying for the statistical parser because it contains errors that are intuitively very clear and resolvable, but which are far beyond the limits of the current statistical tagger. We can take an easy sentence to demonstrate this:

*Je ne le pense pas.*    I do not think so.
*Tu ne le penses pas.*   You do not think so.
*Il ne le pense pas.*    He does not think so.

The verb *pense* is ambiguous[5] in the first person or in the third person. It is usually easy to determine the person just by checking the personal pronoun nearby. For a human or a constraint-based tagger this is an easy task, for a statistical tagger it is not. There are two words between the pronoun and the verb that do not carry any information about the person. The personal pronoun may thus be too far from the verb because bi-gram models can see backward no farther than *le*, and tri-gram models

---

[5] That is not case with all the French verbs, e.g. *Je crois* and *Il croit*.

no farther than *ne le*.

Also, as mentioned earlier, resolving the adjective vs. past participle ambiguity is much harder, if the tagger does not know whether there is an auxiliary verb in the sentence or not.

# 7 Conclusion

We have presented two taggers for french: a statistical one and a constraint-based one.

There are two ways to train the statistical tagger: from a tagged corpus or using a self-organising method that does not need a tagged corpus. We had a strict time limit of one month for doing the tagger and no tagged corpus was available. This is a short time for the manual tagging of a corpus and for the training of the tagger. It would be risky to spend, say, three weeks for writing a corpus, and only one week for training. The size of corpus would have to be limited, because it should be also checked.

We selected the Xerox tagger that learns from an untagged corpus. The task was not as straigthforward as we thought. Without human assistance in the training the result was not impressive, and we had to spend much time tuning the tagger and guiding the learning process. In a month we achieved 95–97 % accuracy.

The training process of a statistical tagger requires some time because the linguistic information has to be incorporated into the tagger one way or another, it cannot be obtained for free starting from null. Because the linguistic information is needed, we decided to encode the information in a more straightforward way, as explicit linguistic disambiguation rules. It has been argued that statistical taggers are superior to rule-based/hand-coded ones because of better accuracy and better adaptability (easy to train). In our experiment, both claims turned out to be wrong.

For the constraint-based tagger we set one month time limit for writing the constraints by hand. We used only linguistic intuition and a very limited set of sentences to write the 75 constraints. We formulated constraints of different accuracy. Some of the constraints are almost 100 % accurate, some of them just describe tendencies.

Finally, when we thought that the rules were good enough, we took two text samples from different sources and tested both the taggers. The constraint-based tagger made several naive errors because we had forgotten, miscoded or ignored some linguistic phenomena, but still, it made only half of the errors that the statistical one made.

A big difference between the taggers is that the tuning of the statistical tagger is very subtle i.e. it is hard to predict the effect of tuning the parameters of the system, whereas the constraint-based tagger is very straightforward to correct.

Our general conclusion is that the hand-coded constraints perform better than the statistical tagger and that we can still refine them. The most important of our findings is that writing constraints that contain more linguistic information than the current statistical model does not take much time.

# References

Jean-Pierre Chanod. Problèmes de robustesse en analyse syntaxique. In *Actes de la conférence Informatique et langue naturelle*. IRIN, Université de Nantes, 1993.

Jean-Pierre Chanod and Pasi Tapanainen. Statistical and Constraint-based Taggers for French. Technical report MLTT-016, Rank Xerox Research Centre, Grenoble, 1994.

Doug Cutting, Julian Kupiec, Jan Pedersen and Penelope Sibun. A Practical Part-of-Speech Tagger. In *Third Conference on Applied Natural Language Processing*. pages 133–140. Trento, 1992.

Ron Kaplan and Martin Kay. Regular Models of Phonological Rule Systems. *Computational Linguistics* Vol. 20, Number 3, pages 331–378.

Fred Karlsson. Constraint Grammar as a Framework for Parsing Running Text. In *proceedings of Coling-90. Papers presented to the 13th International Conference on Computational Linguistics*. Vol. 3, pages 168–173. Helsinki, 1990.

Fred Karlsson, Atro Voutilainen, Juha Heikkilä and Arto Anttila (eds.). *Constraint Grammar: a Language-Independent System for Parsing Unrestricted Text*. Mouton de Gruyter, Berlin, 1995.

Lauri Karttunen. Constructing Lexical Transducers. In *proceedings of Coling-94. The fifteenth International Conference on Computational Linguistics*. Vol I, pages 406–411. Kyoto, 1994.

Lauri Karttunen, Ron Kaplan and Annie Zaenen. Two-level morphology with composition. In *proceedings of Coling-92. The fourteenth International Conference on Computational Linguistics*. Vol I, pages 141–148. Nantes, 1992.

Kimmo Koskenniemi. *Two-level morphology. A general computational model for word-form recognition and production*. University of Helsinki, 1983.

Kimmo Koskenniemi. Finite-state parsing and disambiguation. In *proceedings of Coling-90. Papers presented to the 13th International Conference on Computational Linguistics*. Vol. 2, pages 229–232. Helsinki, 1990.

Kimmo Koskenniemi, Pasi Tapanainen and Atro Voutilainen. Compiling and using finite-state

syntactic rules. In *proceedings of Coling-92. The fourteenth International Conference on Computational Linguistics*. Vol. I, pages 156–162. Nantes, 1992.

Eric Laporte. Experiences in Lexical Disambiguation Using Local Grammars. In *Third International Conference on Computational Lexicography*. pages 163–172. Budapest, 1994.

Emmanuel Roche. Text Disambiguation by finite-state automata, an algorithm and experiments on corpora. In *proceedings of Coling-92. The fourteenth International Conference on Computational Linguistics*. Vol III, pages 993–997. Nantes, 1992.

Emmanuel Roche and Yves Schabes. Deterministic part-of-speech tagging with finite-state transducers. Technical report TR-94-07, Mitsubishi Electric Research Laboratories, Cambridge, USA.

Max Silberztein. *Dictionnaires électroniques et analyse automatique de textes. Le système IN-TEX*. Masson, Paris, 1993.

Pasi Tapanainen and Atro Voutilainen. Tagging accurately – Don't guess if you know. In *Fourth Conference on Applied Natural Language Processing*. pages 47–52. Stuttgart, 1994.

Atro Voutilainen and Pasi Tapanainen. Ambiguity resolution in a reductionistic parser. In *Sixth Conference of the European Chapter of the ACL*. pages 394–403. Utrecht, 1993.

# A  The restricted tag set

In this appendix the tag set is represented. Besides the following tags, there may also be some word-specific tags like *PREP-DE*, which is the preposition reading for words *de*, *des* and *du*, i.e. word *de* is initially ambiguous between *PREP-DE* and *PC*. This information is mainly for the statistical tagger to deal with, for instance, different prepositions in a different way. The constraint-based tagger does not need this because it has direct access to word forms anyway. After disambiguation, the word-specific tags may be cleaned. The tag *PREP-DE* is changed back into *PREP*, to reduce the redundant information.

- **DET-SG**: Singular determiner e.g. *le, la, mon, ma*. This covers masculine as well as feminine forms. Sample sentence: *Le chien dort dans la cuisine*. (The dog is sleeping in the kitchen).

- **DET-PL** Plural determiner e.g. *les, mes*. This covers masculine as well as feminine forms. Sample sentence: *Les enfants jouent avec mes livres*. (The children are playing with my books.)

- **ADJ-INV** Adjective invariant in number e.g. *heureux*. Sample sentence: *Le chien est heureux quand les enfants sont heureux*. (The dog is happy when the children are happy.)

- **ADJ-SG** Singular adjective e.g. *gentil, gentille*. This covers masculine as well as feminine forms. Sample sentence: *Le chien est gentil*. (The dog is nice.)

- **ADJ-PL** Plural adjective e.g. *gentils, gentilles*. This covers masculine as well as feminine forms. Sample sentence: *Ces chiens sont gentils*. (These dogs are nice.)

- **NOUN-INV** Noun invariant in number e.g. *souris, Français*. This covers masculine as well as feminine forms. Sample sentence: *Les souris dansent*. (The mice are dancing.)

- **NOUN-SG** Singular noun e.g. *chien, fleur*. This covers masculine as well as feminine forms. Sample sentence: *C'est une jolie fleur*. (It is a nice flower.)

- **NOUN-PL** Plural noun e.g. *chiens, fleurs*. This covers masculine as well as feminine forms. Sample sentence: *Nous aimons les fleurs*. (We like flowers.)

- **VAUX-INF** Auxiliary verb, infinitive *être, avoir*. Sample sentence: *Le chien vient d'être puni*. (The dog has just been punished.)

- **VAUX-PRP** Auxiliary verb, present participle *étant, ayant*.

- **VAUX-PAP** Auxiliary verb, past participle e.g. *été, eu*. Sample sentence: *Le théorème a été démontré*. (The theorem has been proved.)

- **VAUX-P1P2** Auxiliary verb, covers any 1st or 2nd person form, regardless of number, tense or mood, e.g. 1st person singular present indicative, 2nd person plural imperative: *ai, soyons, es*. Sample sentence: *Tu es fort*. (You are strong.)

- **VAUX-P3SG** Auxiliary verb, covers any 3rd person singular form e.g. *avait, sera, es*. Sample sentence: *Elle est forte*. (She is strong.)

- **VAUX-P3PL** Auxiliary verb, covers any 3rd person plural form e.g. *ont, seront, avaient*. Sample sentence: *Elles avaient dormi*. (They had slept.)

- **VERB-INF** Infinitive verb e.g. *danser, finir, dormir*. Sample sentence: *Le chien aime dormir*. (The dog enjoys sleeping.)

- **VERB-PRP** Present participle e.g. *dansant, finissant, aboyant*. Sample sentence: *Le chien arrive en aboyant*. (The dog is coming and it is barking.)

- **VERB-P1P2** Any 1st or 2nd person verb form, regardless of number, tense or mood e.g. 1st person singular present indicative,

2nd pers plural imperative: *chante, finissons*. Sample sentence: *Je chante.* (I sing.)

- **VERB-P3SG** Any 3rd person singular verb form e.g. *chantera, finit, aboie*. Sample sentence: *Le chien aboie.* (The dog is barking.)
- **VERB-P3PL** Any 3rd person plural verb form e.g. *chanteront, finissent, aboient*. Sample sentence: *Les chiens aboient.* (The dogs are barking.)
- **PAP-INV** Past participle invariant in number e.g. *surpris*. Sample sentence: *Le chien m'a surpris.* (The dog surprised me.)
- **PAP-SG** Singular past participle e.g. *fini, finie*. This covers masculine as well as feminine forms. Sample sentence: *La journée est finie.* (The day is over.)
- **PAP-PL** Plural past participle e.g. *finis, finies*. This covers masculine as well as feminine forms. Sample sentence: *Les travaux sont finis.* (The work is finished.)
- **PC** Non-nominative clitic pronoun such as *me, le*. Sample sentence: *Il me l'a donné.* (He gave it to me.)
- **PRON** 3rd person pronoun, relative pronouns excluded. e.g. *il, elles, chacun*. Sample sentence: *Il a parle à chacun.* (He spoke to every person.)
- **PRON-P1P2** 1st or 2nd person pronoun e.g. *je, tu, nous*. Sample sentence: *Est-ce que tu viendras avec moi?* (Will you come with me?)
- **VOICILA** Reserved for words *voici* and *voilà*. Sample sentence: *Voici mon chien.* (Here is my dog.)
- **ADV** Adverbs e.g. *finalement*. Sample sentence: *Le jour est finalement arrivé.* (The day has finally come.)
- **NEG** Negation particle. Reserved for the word *ne*. Sample sentence: *Le chien ne dort pas.* (The dog is not sleeping.)
- **PREP** Preposition e.g. *dans*. Sample sentence: *Le chien dort dans la cuisine.* (The dog sleeps in the kitchen.)

  For statistical taggers this group may be divided into subgroups for different preposition groups, like *PREP-DE*, *PREP-A*, etc.

- **CONN** Connector. This class includes coordinating conjuctions such as *et*, subordinate conjunctions such as *lorsque*, relative or interrogative pronouns such as *lequel*. Words like *comme* or *que* which have very special behaviour are not coded as *CONN*. Sample sentence: *Le chien et le chat dorment quand il pleut.* (The dog and the cat sleep when it rains.)

  For statistical taggers this group may be divided into subgroups for different connectors, like *CONN-ET*, *CONN-Q*, etc.

- **COMME** Reserved for all instances of the word *comme*. Sample sentence: *Il joue comme un enfant.* (He plays like a child.)
- **CONJQUE** Reserved for all instances of the word *que*.
- **NUM** Numeral e.g. 12,7, 120/98, 34+0.7.
- **HEURE** String representing time e.g. 12h24, 12:45:00.
- **MISC** Miscellaneous words, such as: interjection *oh*, salutation *bonjour*, onomatopoeia *miaou*, wordparts i.e. words that only exist as part of a multi-word expression, such as *priori*, as part of *a priori*.
- **CM** Comma.
- **PUNCT** Punctuation other than comma.